\documentclass[11pt]{article}

\usepackage[T1]{fontenc}
\usepackage{hyperref}
\usepackage{url}
\usepackage{graphicx} 
\usepackage{subfigure} 
\usepackage{bm} 
\usepackage{amsmath}
\usepackage{natbib} 
\usepackage{multirow}

\usepackage{algorithm}
\usepackage{algorithmic}

\usepackage[a4paper, total={6in, 8in}]{geometry}

\title{Online Meta-learning by Parallel Algorithm Competition}

\author{
Stefan Elfwing$^{\rm a}$ \\
\texttt{elfwing@atr.jp}
\and
Eiji Uchibe$^{\rm a,b}$ \\
\texttt{uchibe@atr.jp} 
\and
Kenji Doya$^{\rm b}$ \\
\texttt{doya@oist.jp} 
}

\date{$^{\rm a}$Dept. of Brain Robot Interface, ATR Computational
  Neuroscience Laboratories, 2-2-2 Hikaridai, Seikacho, Soraku-gun,
  Kyoto 619-0288, Japan\\ 
  $^{\rm b}$Neural Computation Unit, Okinawa
  Institute of Science and Technology Graduate University, 1919-1
  Tancha, Onna-son, Okinawa 904-0495, Japan }

\begin{document}

\maketitle

\begin{abstract}
\noindent The efficiency of reinforcement learning algorithms depends
critically on a few meta-parameters that modulates the learning
updates and the trade-off between exploration and exploitation. The
adaptation of the meta-parameters is an open question in reinforcement
learning, which arguably has become more of an issue recently with the
success of deep reinforcement learning in high-dimensional state
spaces. The long learning times in domains such as Atari 2600 video
games makes it not feasible to perform comprehensive searches of
appropriate meta-parameter values. We propose the Online Meta-learning
by Parallel Algorithm Competition (OMPAC) method. In the OMPAC method,
several instances of a reinforcement learning algorithm are run in
parallel with small differences in the initial values of the
meta-parameters. After a fixed number of episodes, the instances are
selected based on their performance in the task at hand. Before
continuing the learning, Gaussian noise is added to the meta-parameters
with a predefined probability. We validate the OMPAC method by
improving the state-of-the-art results in stochastic SZ-Tetris and in
standard Tetris with a smaller, 10$\times$10, board, by $31\,\%$ and
$84\,\%$, respectively, and by improving the results for deep
Sarsa($\lambda$) agents in three Atari 2600 games by $62\,\%$ or
more. The experiments also show the ability of the OMPAC method to
adapt the meta-parameters according to the learning progress in
different tasks.
\end{abstract}

\section{Introduction}
The efficiency of reinforcement learning~\citep{Sutton98} algorithms
depends critically on a few meta-parameters that modulates the
learning updates and the trade-off between exploration for new
knowledge and exploitation of existing knowledge. Ideally, these
meta-parameters should not be fixed during learning. Instead, they
should be adapted according to the current learning progress in the
task at hand. The adaptation of the meta-parameters is an open
question in reinforcement learning, which arguably has become more of
an issue recently with the success of deep reinforcement learning in
tasks with high-dimensional state spaces, such as ability of the DQN
algorithm to achieve human-level performance in many Atari 2600 video
games~\citep{Mnih15}. The complexity of and the long learning times in
such tasks, where the episode length often increases with improvements
in performance, makes it not feasible to perform comprehensive
grid-like searches of appropriate meta-parameter values.

We propose the Online Meta-learning by Parallel Algorithm Competition
(OMPAC) method. The idea behind OMPAC is simple. Run several instances
of a reinforcement algorithm in parallel, with small differences in
the initial values of the meta-parameters. After a fixed number of
episodes, the instances are selected based on their performance in the
task at hand. Before continuing the learning, Gaussian noise is added
to the meta-parameters with a predefined probability. The OMPAC method
is similar to a Lamarckian~\citep{Lamarck1809} evolutionary process
without the crossover operator, but with two main differences compared
with standard applications of artificial evolution. First, the goal is
not to find the parameters that represent the optimal solutions
directly, instead the goal is to find the meta-parameters that enable
reinforcement learning agents to learn more efficiently. Second, the
goal is not to find the best set of fixed parameters, instead the goal
is to adapt the values of the meta-parameters according to the current
learning progress. Importantly, the OMPAC method differentiates
between the objective of the learning (i.e., maximize the expected
accumulated discounted rewards, in the cases of value-based
reinforcement learning algorithms used in this study) and the overall
goal of the task which is used as the selection criteria for continued
learning.

The studies most related to our research have used a Darwinian
evolutionary approach (i.e., the learning started from scratch in each
generation) to find appropriate fixed values of the
meta-parameters~\citep{Unemi94,Eriksson03,Elfwing08,Elfwing11}.
\citet{Schweighofer03} proposed a meta-learning method based on the
mid-term and the long-term running averages of the reward, and
~\citet{Kobayashi09} proposed a meta-learning method based on the
absolute values of the TD-errors. Both methods require the setting of
meta-meta-parameters, which is a non-trivial task. Proposed approaches
for adapting individual meta-parameters include the incremental delta
bar delta method~\citep{Sutton92a} to tune $\alpha$, a method based on
variance of the action value function~\citep{Ishii02} to tune the
inverse temperature $\beta$ in softmax action selection, and a
Bayesian model averaging approach~\citep{Downey10} and the Adaptive
$\lambda$ Least-Squares Temporal Difference Learning
method~\citep{Mann16} to tune $\lambda$. \citet{FranoisLavet15}
demonstrated that the performance of DQN could be improved in some
Atari 2600 games by a rather ad-hoc tuning scheme that increase
$\gamma$ ($\gamma \gets \min(0.02+0.98\gamma,0.99)$) and decrease
$\alpha$ ($\alpha \gets 0.98\alpha$).

We validate the OMPAC method in stochastic SZ-Tetris and in standard
Tetris with a smaller, 10$\times$10, board, and in the Atari 2600
domain. To be able to directly compare the learning performance with
and without the OMPAC method, we use the same experimental setups as
in our earlier study \citep{Elfwing17}, where we proposed the
sigmoid-weighted linear (SiL) unit and its derivative function (dSiL)
as activation functions for neural network function approximation in
reinforcement learning. In both SZ-Tetris and 10$\times$10 Tetris, we
train shallow neural network agents with dSiL units in the hidden
layer and improve the state-of-the-art scores by $31\,\%$ and
$84\,\%$, respectively. In three Atari 2600 games, we train deep
neural network agents with SiL units in the convolutional layers and
dSiL units in the hidden fully-connected layer and improve the
performance by $62\,\%$ or more. Finally, we demonstrate the ability
of the OMPAC method to achieve efficient learning even if the initial
values of the meta-parameters are not suitable for the task at hand.

\section{Background}
\subsection[TD(lambda) and Sarsa(lambda)]{TD($\lambda$) and Sarsa($\lambda$)}
In this study, we use two reinforcement learning algorithms:
TD($\lambda$)~\citep{Sutton88} and
Sarsa($\lambda$)~\citep{Rummery94,Sutton96}. TD($\lambda$) learns an
estimate of the state-value function, $V^{^{\pi}}$, and
Sarsa($\lambda$) learns an estimate of the action-value function,
$Q^{^{\pi}}$, while the agent follows policy $\pi$. If the
approximated value functions, $V_t\approx V^{^{\pi}}$ and $Q_t\approx
Q^{^{\pi}}$, are parameterized by the parameter vector
$\boldsymbol{\theta}_t$, then the gradient-descent learning update of
the parameters is computed by
\begin{equation}
  \boldsymbol{\theta}_{t+1} = \boldsymbol{\theta}_t + \alpha\delta_t\boldsymbol{e}_t,
  \label{eq:upd}
\end{equation}
where the TD-error, $\delta_t$, is
\begin{equation}
\delta_t = r_{t} +\gamma V_t(s_{t+1}) - V_t(s_{t})
\end{equation}
for TD($\lambda$) and 
\begin{equation}
\delta_t = r_{t} +\gamma Q_t(s_{t+1},a_{t+1}) - Q_t(s_{t},a_{t})
\end{equation}
for Sarsa($\lambda$). The eligibility trace vector,
$\boldsymbol{e}_t$, is
\begin{equation}
\boldsymbol{e}_t = \gamma\lambda \boldsymbol{e}_{t-1} + \nabla_{\boldsymbol{\theta}_t} V_t(s_{t}), \ \boldsymbol{e}_0 = \boldsymbol{0}, 
\label{eq:tracesV}
\end{equation}
for TD($\lambda$) and
\begin{equation}
\boldsymbol{e}_t = \gamma\lambda \boldsymbol{e}_{t-1} + \nabla_{\boldsymbol{\theta}_t} Q_t(s_{t},a_{t}), \ \boldsymbol{e}_0 = \boldsymbol{0},
\label{eq:tracesQ}
\end{equation}
for Sarsa($\lambda$). Here, $s_t$ is the state at time $t$, $a_t$ is
the action selected at time $t$, $r_t$ is the reward for taking action
$a_t$ in state $s_t$, $\alpha$ is the learning rate, $\gamma$ is
the discount factor of future rewards, $\lambda$ is the trace-decay
rate, and $\nabla_{\boldsymbol{\theta}_t} V_t$ and
$\nabla_{\boldsymbol{\theta}_t} Q_t$ are the vectors of partial
derivatives of the function approximators with respect to each
component of $\boldsymbol{\theta}_t$.

\subsection{Sigmoid-weighted Linear Units}
We recently proposed~\citep{Elfwing17} the sigmoid-weighted linear (SiL)
unit and its derivative function (dSiL) as activation functions for
neural network function approximation in reinforcement learning. The
activation $a_k$ of SiL unit $k$ for an input vector $\mathbf{s}$ is
computed by the sigmoid function, $\sigma(\cdot)$, multiplied by its
input, $z_k$:
\begin{eqnarray}
a_k(\mathbf{s}) &=& z_k\sigma(z_k), \\
z_k(\mathbf{s}) &=& \sum_iw_{ik}s_i + b_k, \\
\sigma(x) &=& \frac{1}{1+e^{-x}}. 
\end{eqnarray}
Here, $w_{ik}$ is the weight connecting state $s_i$ and hidden unit $k$,
$b_k$ is the bias weight for hidden unit $k$. The activation of the dSiL
unit is computed by:
\begin{equation}
a_k(\mathbf{s}) = \sigma(z_k)\left(1 + z_k(1 - \sigma(z_k))\right).
\end{equation}

The derivative of the activation function of the SiL unit, used for
gradient-descent learning updates of the neural network weight
parameters (see Equations \ref{eq:tracesV} and \ref{eq:tracesQ}), is
given by
\begin{equation}
\nabla_{w_{ik}} a_k(\mathbf{s}) = \sigma(z_k)\left(1 + z_{k}(1 - \sigma(z_k))\right)s_i,
\end{equation}
and the derivative of the activation function of the dSiL unit is given by
\begin{eqnarray}
\nabla_{w_{ik}} a_k(\mathbf{s}) &=& \sigma(z_k)(1-\sigma(z_k))(2 + \nonumber \\
 {} & {} & z_{k}(1 - \sigma(z_k))- z_{k}\sigma(z_k))s_i.
\end{eqnarray}

\subsection{Action selection}
We use softmax action selection with a Boltzmann distribution in all
experiments. For Sarsa($\lambda$), the probability to select action
$a$ in state $s$ is defined as
\begin{equation}
\pi(a | s) = \frac{\exp({Q(s,a)/\tau})}{\sum_b \exp({Q(s,b)/\tau})}.
\end{equation}
For the model-based TD($\lambda$) algorithm, we select an action $a$
in state $s$ that leads to the next state $s'$ with a the probability
defined as
\begin{equation}
\pi(a | s) = \frac{\exp({V(f(s,a))/\tau})}{\sum_{b} \exp({V(f(s,b))/\tau})}.
\end{equation}
Here, $f(s,a)$ returns the next state $s'$ according to the
state transition dynamics and $\tau$ is the temperature
that controls the trade-off between exploration and exploitation. We
used hyperbolic discounting of the temperature and the temperature was
decreased after every episode $i$:
\begin{equation}
\tau(i) = \frac{\tau_0}{1 + \tau_ki}.
\label{eq:tau}
\end{equation}
Here, $\tau_0$ is the initial temperature and $\tau_k$ controls the
rate of discounting.

\section{The OMPAC method}
In this section, we present the OMPAC (Online Meta-learning by Parallel
Algorithm Competition) method. Algorithm~\ref{alg:OMPAC} shows the
pseudo-code for the OMPAC method with Sarsa($\lambda$) and softmax
action selection. Several, $N$, instances of the algorithm are run in
parallel, with small differences in meta-parameter values. After a
fixed number of episodes, the instances are selected for continued
learning based on their performance in the task. The OMPAC method
differentiates between the goal of the learning (i.e., maximize the
expected accumulated discounted rewards, in the cases of
Sarsa($\lambda$)) and the overall goal of the task as measured by the
score, $F_i$, for each instance $i$ in each generation. In this study,
the score is equal to the total number of points scored in the games
we consider.

\begin{algorithm*}[!htb]
\caption{OMPAC with Sarsa($\lambda$) and softmax action selection}
\label{alg:OMPAC} 
\begin{algorithmic}
  \STATE Initialize matrix of parameter vectors, $\boldsymbol{\Theta} = [\boldsymbol{\theta}_1, \dots, \boldsymbol{\theta}_i, \dots, \boldsymbol{\theta}_N]$.
  \STATE Initialize matrix of meta-parameter vectors, $\boldsymbol{\Psi} = [\boldsymbol{\psi}_1, \dots, \boldsymbol{\psi}_i, \dots, \boldsymbol{\psi}_N]$.
  \\
  \FOR{\textbf{each} generation}
    \STATE Initialize vector of scores $\boldsymbol{F} \gets \boldsymbol{0}$
    \FORALL[Parallel for loop]{$i = 1$ \TO $N$}
      \FOR{\textbf{each} episode of the generation}
        \STATE Get initial state $s_i$ and select action $a_i\sim \pi(\cdot|s_i)$. \COMMENT{Softmax action selection}
        \STATE $\boldsymbol{e_i} \gets \boldsymbol{0}$ 
        \WHILE{$s_i$ is not terminal}
          \STATE Take action $a_i$, observe reward $r_i$, score $f_i$, and next state $s'_i$.
            \STATE $F_i \gets F_i + f_i$.

            \STATE $\boldsymbol{e}_i \gets \gamma_i\lambda_i \boldsymbol{e}_i + \nabla_{\boldsymbol{\theta}_i} Q(s_i,a_i|\boldsymbol{\theta}_i)$
            \IF{$s'_i$ is terminal}
              \STATE $\delta_i \gets r_i - Q(s_i,a_i|\boldsymbol{\theta_i})$ 
            \ELSE
              \STATE Select new action $a'_i\sim \pi(\cdot|s'_i)$.
              \STATE $\delta_i \gets r_i  + \gamma_iQ(s'_i,a'_i|\boldsymbol{\theta_i}) - Q(s_i,a_i|\boldsymbol{\theta_i})$
              \STATE $a_i \gets a'_i$  
            \ENDIF
            \STATE $\boldsymbol{\theta}_i \gets \boldsymbol{\theta}_i + \alpha_i\delta_i\boldsymbol{e}_i$
            \STATE $s_i \gets s'_i$
          \ENDWHILE
       \ENDFOR
    \ENDFOR
    \STATE $[\boldsymbol{\Theta}, \boldsymbol{\Psi}] \gets$ \textsc{Selection}($\boldsymbol{F}, \boldsymbol{\Theta}, \boldsymbol{\Psi}$) 
    \STATE $\boldsymbol{\Psi} \gets$ \textsc{AddNoise}($\boldsymbol{\Psi}$)
  \ENDFOR
\end{algorithmic}
\end{algorithm*}

We use stochastic universal sampling~\citep[SUS;][]{Baker87} combined
with elitism as the \textsc{Selection}() method. SUS is a fitness
proportionate selection method with no bias and minimal
variance. Instead of spinning an imaginary roulette wheel with one
pointer $N$ times, as in roulette wheel selection, SUS spins the wheel
once with $N$ equally-spaced pointers. We first select the best
algorithm instance and it continues learning without adding noise to
any of the meta-parameters (i.e., elitism). We then use SUS to select the
remaining instances and we add Gaussian noise to the meta-parameters with
a predefined probability $p_n$. To ensure that changes in
meta-parameter values are not too large and therefore disruptive to
the learning process, we use a \textsc{AddNoise}() method that adds
noise, $\epsilon$, to a meta-parameter, $\psi$, drawn from a normal
distribution with standard deviation that depends on the magnitude of
meta-parameter multiplied by predefined factor $\eta_n$. If $0 \le
\psi \le 1$ ($\alpha$, $\gamma$, and $\lambda$ in this study), then:
\begin{equation}
\epsilon \sim 
\begin{cases}
\mathcal{N}\left(0, \left(\psi\eta_n\right)^2  \right) & if \psi \le 0.5,\\  
\mathcal{N}\left(0, \left(\left(1-\psi\right)\eta_n\right)^2  \right) & if \psi > 0.5, 
\label{eq:addN1}
\end{cases}
\end{equation}
otherwise ($\tau_0$ and $\tau_k$ in this study):
\begin{equation}
\epsilon \sim 
\mathcal{N}\left(0, \left(\psi\eta_n\right)^2  \right).  
\label{eq:addN2}
\end{equation}
We also use Equations~\ref{eq:addN1} and \ref{eq:addN2} to initialize
the meta-parameters by adding Gaussian noise to common starting values.

\section{Experiments}
To be able to directly compare the learning performance with and
without the OMPAC method, we used the same experimental setups as in
our earlier study~\citep{Elfwing17}. In stochastic SZ-Tetris and
standard Tetris with a smaller, 10$\times$10, board size, we trained
agents with shallow neural network function approximators with dSiL
hidden units using TD($\lambda$) and softmax action selection
(hereafter, \emph{shallow dSiL agents}). In the Atari 2600 domain, we
trained agents with deep convolutional neural network function
approximators with SiL hidden units in the convolutional layers and
dSiL hidden units in the fully-connected layer using Sarsa($\lambda$)
and softmax action selection (hereafter, \emph{deep SiL agents}). In
all OMPAC experiments, the number of parallel algorithm instances $N$
was set to $12$, the number of learning episodes in each generation
was set to 100, the probability of adding Gaussian noise to a
meta-parameter $p_n$ was set to $0.1$, and the factor controlling the
magnitude of the Gaussian noise $\eta_n$ was set to $0.05$. The values
were determined by preliminary experiments in stochastic SZ-Tetris.
 
\subsection{Tetris}
Due to prohibitively long learning times (in the case of high
performance), it is not feasible to apply value-based reinforcement
learning to standard Tetris with a board height of 20. The current
state-of-the-art result for a single run of an algorithm, achieved
by the CBMPI algorithm~\citep{Gabillon13}, is a mean score of 51
million cleared lines. We instead consider stochastic
SZ-Tetris~\citep{Burgiel97,Szita10}, which only uses the S-shaped and
the Z-shaped tetrominos, and standard Tetris with a smaller, 10$\times$10,
board.

In both versions of Tetris, in each time step, a randomly selected
tetromino appears above the board. The agent selects a rotation and a
horizontal position, and the tetromino drops down the board, stopping
when it hits another tetromino or the bottom of the board. If a row is
completed, then it disappears. An episode ends when a tetromino does
not fit within the board. The agent gets a score equal to the number
of cleared lines, with a maximum of 2 points in SZ-Tetris and of 4 points
in 10$\times$10 Tetris (only achievable by the stick-shaped tetromino). The
possible number of actions is 17 for the S-shaped and the Z-shaped
tetrominos. For the additional five tetrominos in 10$\times$10 Tetris, the
possible numbers of actions are 9 for the block-shaped tetromino, 17
for stick-shaped tetromino, and 34 for the J-, L- and T-shaped
tetrominos.

We recently achieved the current state-of-the-art results~\citep{Elfwing17} using
shallow dSiL agents. In SZ-Tetris, a shallow dSiL
agent with 50 hidden nodes achieved a final (over 1,000 episodes) mean
score of 263 points when averaged over 10 separate runs and of 320
points for the best run. In 10$\times$10 Tetris, a shallow dSiL agent
with 250 hidden nodes achieved a final (over 10,000 episodes) mean
score of 4,900 points when averaged over 5 separate runs of and 5,300
points for the best run, which improved the average score of 4,200
points and the best score of 5,000 points achieved by the CBMPI
algorithm~\citep{Gabillon13}.

Using the OMPAC method we trained shallow dSiL agents with 50 hidden
units in SZ-Tetris and 250 hidden units in 10$\times$10 Tetris.  The
features were similar to the original 21 features proposed
by~\citet{Bertsekas96}, except for not including the maximum column
height and using the differences in column heights instead of the
absolute differences. The binary state vectors were of length $460$ in
SZ-Tetris and $260$ in 10$\times$10 Tetris.
We used the following reward function proposed by \citet{Fausser13}:
\begin{equation}
r(\boldsymbol{s}) = e^{-(\textrm{number of holes in }\boldsymbol{s})/z}.
\label{eq:tetris_r}
\end{equation}
Here, $z$ was set to $33$ for SZ-Tetris and $33/2$ for 10$\times$10
Tetris. We used the meta-parameters in \citet{Elfwing17} as starting
values to initialize the meta-parameters according to
Equations~\ref{eq:addN1} and \ref{eq:addN2}: $\alpha$: 0.001,
$\gamma$: 0.99, $\lambda$: 0.55, $\tau_0$: 0.5, and $\tau_k$:
0.00025. The experiments ran for 2,000 generations (i.e., 200,000
episodes of learning in total) in SZ-Tetris and 2,500 generations
(250,000 episodes) in 10$\times$10 Tetris. The score used for
selection was the total number of points received by an algorithm
instance in one generation.
 
\begin{figure}[!htb]
   \begin{center}
    \subfigure
    {
      \includegraphics[width=0.47\columnwidth]{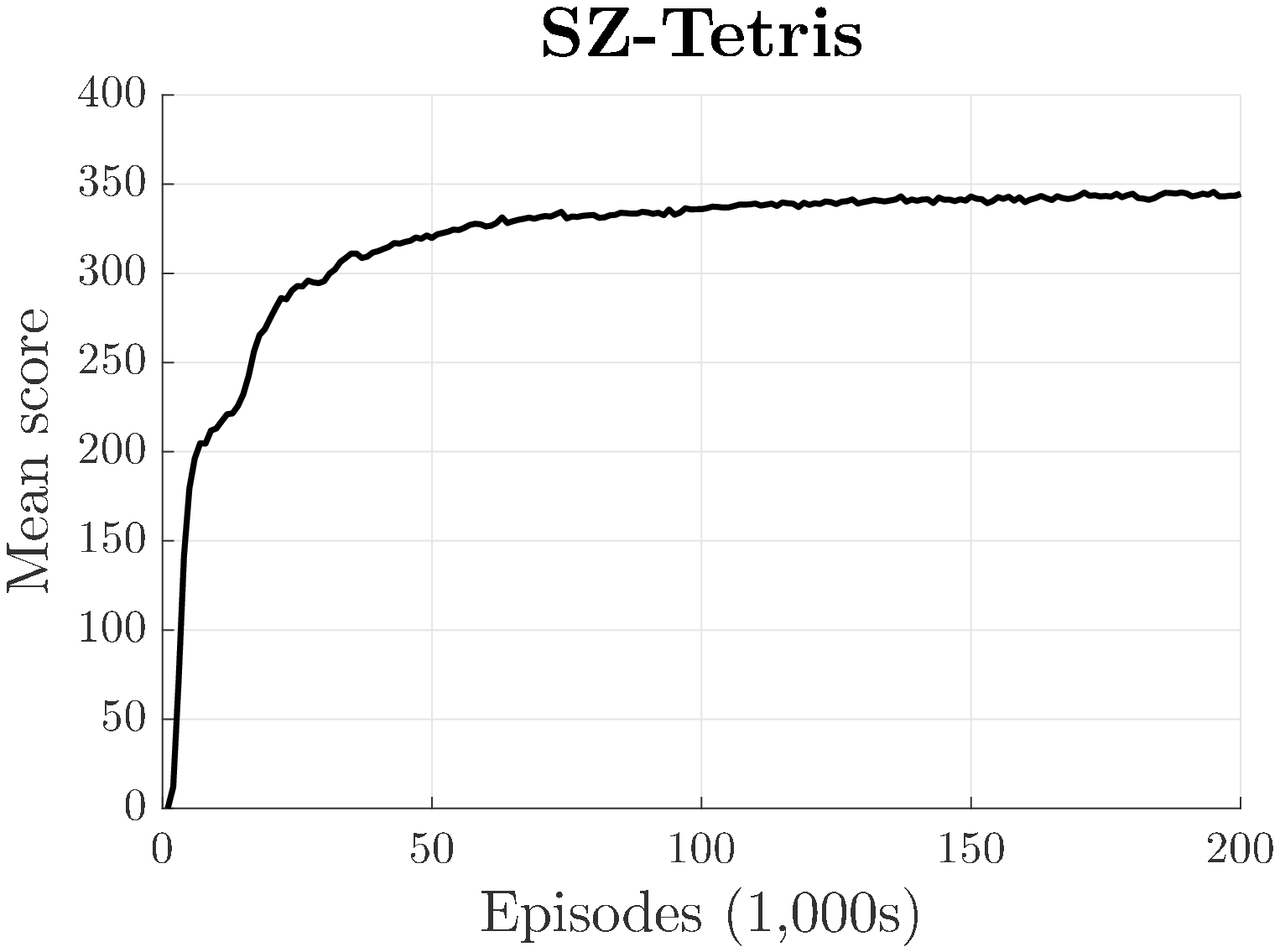}
    }
    \subfigure
    {
      \includegraphics[width=0.47\columnwidth]{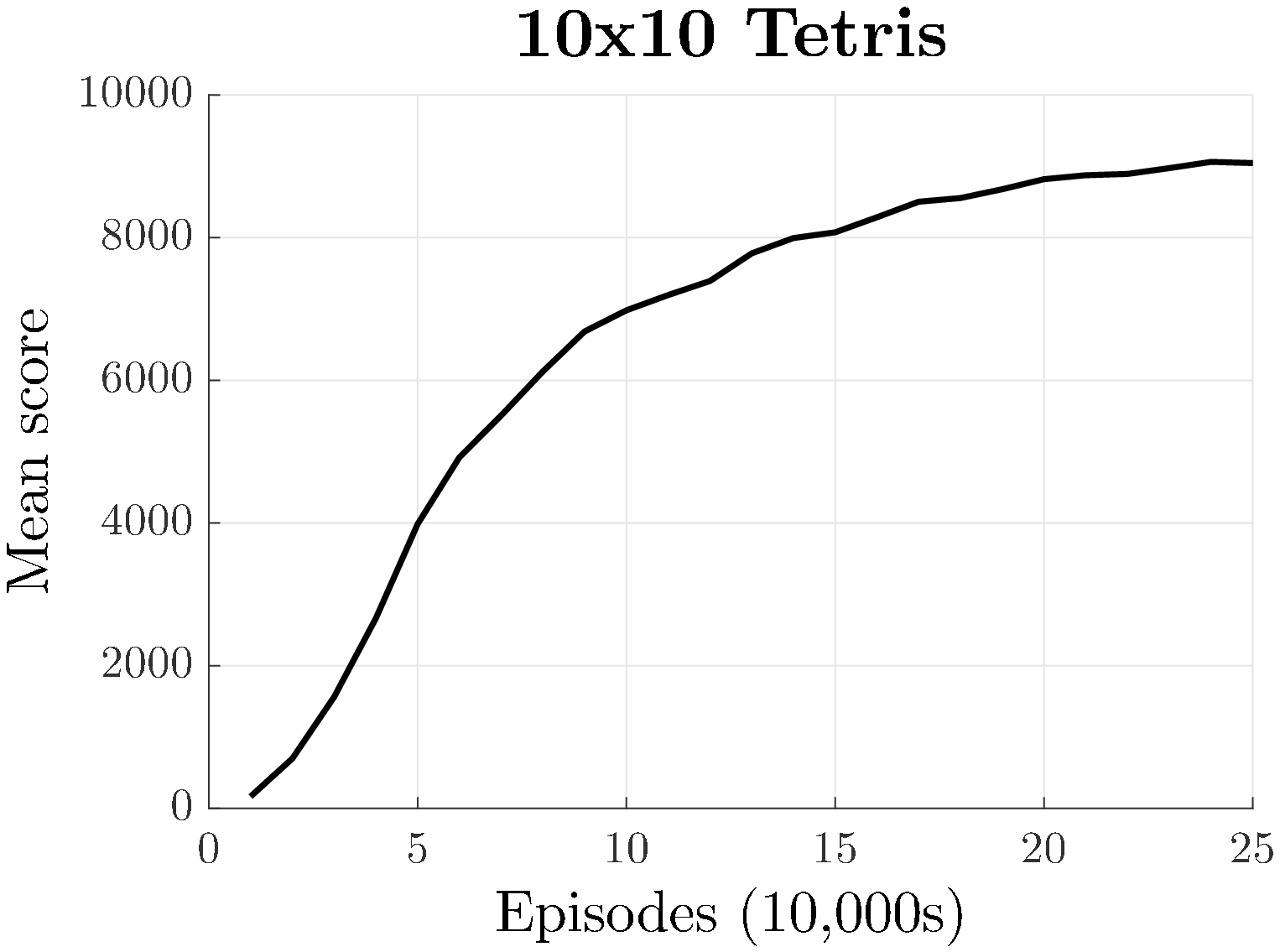} 
    }
  \end{center}
  \caption{Average learning curves over the 12 algorithm instances in
    SZ-Tetris and in 10$\times$10 Tetris.}
  \label{fig:res_tet} 
\end{figure}
Figure~\ref{fig:res_tet} shows the average learning curves for the
OMPAC method in SZ-Tetris and 10$\times$10 Tetris. The scores were
computed as mean scores over every 10 generations (1,000 episodes) in
SZ-Tetris and every 100 generations (10,000 episodes) in 10$\times$10
Tetris. The final scores are large and significant improvements of the
previous state-of-the-art results achieved by shallow dSiL agents
without OMPAC adaptation. The final average score of $345$ points is a
$82$ points or $31\,\%$ improvement in SZ-Tetris, and the final
average score of 9,000 points is a 4,100 points or $84\,\%$
improvement in 10$\times$10 Tetris.

\begin{figure}[!htb]
   \begin{center}
     \subfigure {
       \includegraphics[width=0.47\columnwidth]{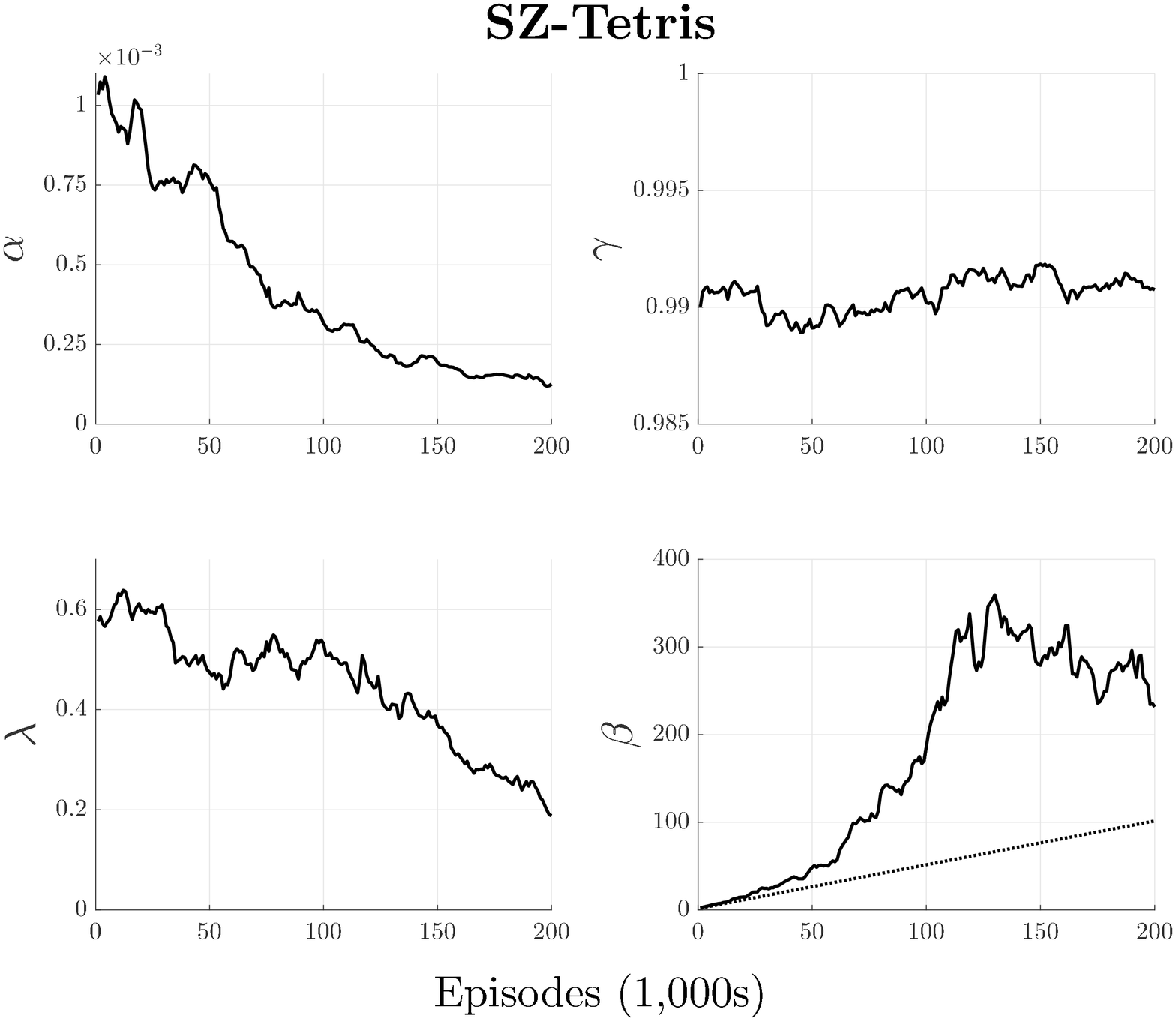}
     } \subfigure {
       \includegraphics[width=0.47\columnwidth]{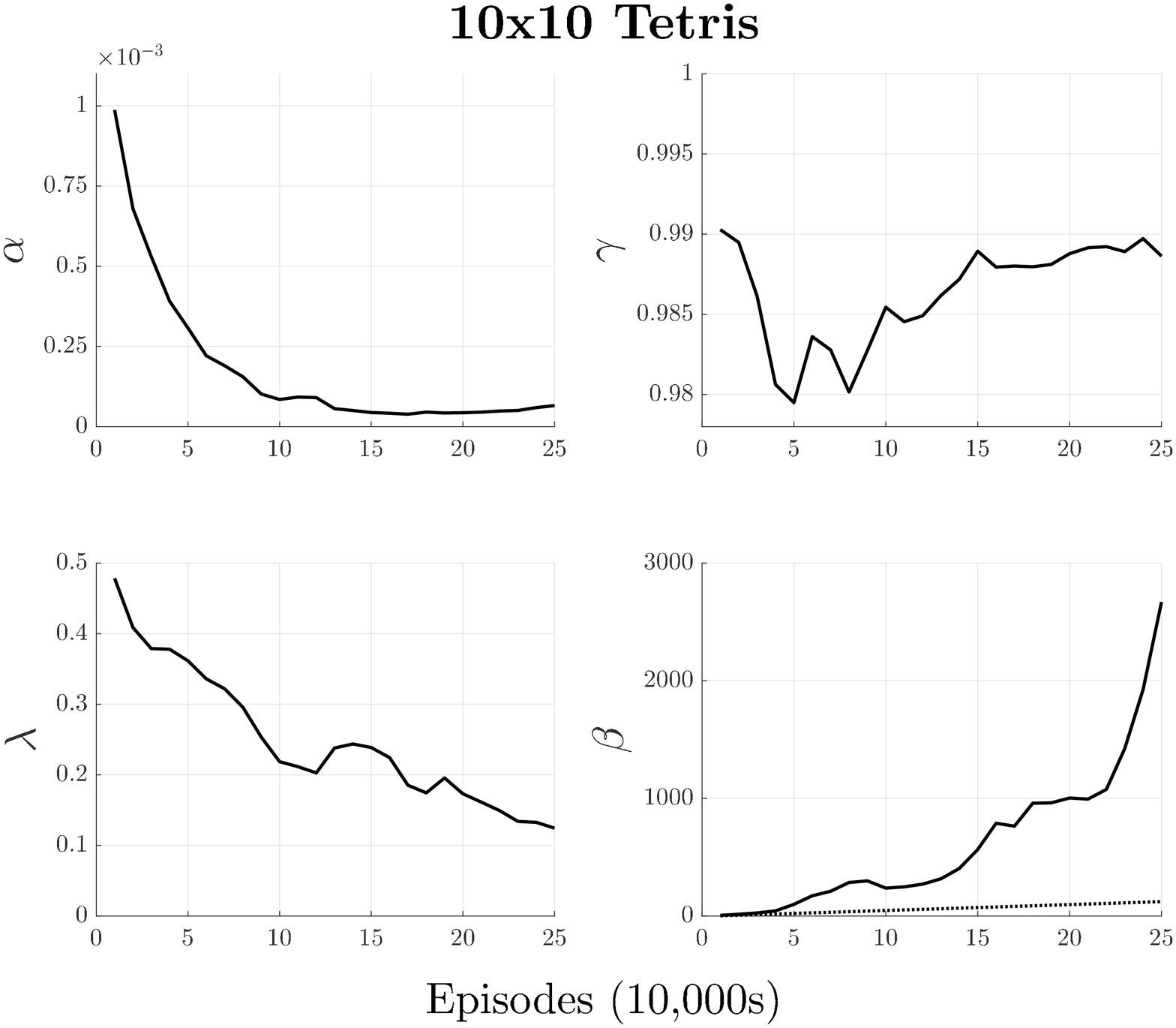}
     }
  \end{center}
  \caption{Average values of the meta-parameters in SZ-Tetris and
    10$\times$10 Tetris for OMPAC adaptation. For visualization
    purposes, instead of showing $\tau(i)$, the figure shows the
    inverse temperature $\beta(i) = 1/\tau(i) =
    (1+\tau_ki)/\tau_0$. The dotted lines in the $\beta$-panels show
    the $\beta(i)$-values for the starting values of $\tau_0$ and
    $\tau_k$.}
  \label{fig:meta_tet}
\end{figure}

Figure~\ref{fig:meta_tet} shows the average values of the
meta-parameters computed over the 12 algorithm instances. 
The adaptations of $\alpha$ and $\lambda$ were similar in both
games. The values of $\alpha$ decreased by about an order of
magnitude, from $1.0{\times}10^{-3}$ to $1.25{\times}10^{-4}$ and
$6.57{\times}10^{-5}$, and the values of $\lambda$ decreased from 0.55
to 0.188 and 0.125, in SZ-Tetris and in 10$\times$10 Tetris,
respectively. More interestingly, while the value of $\gamma$ in
SZ-Tetris was relatively stable around 0.99 (final average value of
0.9908), the value of $\gamma$ in 10$\times$10 Tetris first decreased to
about 0.98 and then increased back towards 0.99 (final average value
of 0.9886). The value of the inverse temperature $\beta = 1/\tau$ in
SZ-Tetris increased, in tandem with the improvement in mean score, to
over 300 during first $\sim$125,000 episodes. During the last $\sim$75,000
episodes, there was only a very small ($\sim$5 points) improvement in
mean score and the value of $\beta$ decreased slowly, reaching a final
average value of 232. In 10$\times$10 Tetris, the mean score improved
continuously and the value of $\beta$ increased over the whole
learning process, reaching a final average value of $2670$ (i.e.,
approximately greedy action selection).

\subsection{Atari 2600 games}
We evaluated the OMPAC method in the Atari 2600 domain using the
Arcade Learning Environment~\citep{Bellemare13}. We followed the
experimental setup in our earlier study~\citep{Elfwing17}. The raw
210$\times$160 Atari 2600 RGB frames were pre-processed by extracting
the luminance channel, taking the maximum pixel values over
consecutive frames to prevent flickering, and then downsampling the
grayscale images to 105$\times$80. The deep convolutional neural
network used by the deep SiL agent consisted of two convolutional
layers with SiL units (16 filters of size 8$\times$8 with a stride of
4 and 32 filters of size 4$\times$4 with a stride of 2), each followed
by a max-pooling layer (pooling windows of size 3$\times$3 with a
stride of 2), a fully-connected hidden layer with 512 dSiL units, and
a fully-connected linear output layer with 4 to 18 output (or
action-value) units, depending on the number of valid actions in the
considered game. We used frame skipping where actions were selected
every fourth frame and repeated for the next four frames. The input to
the network was a 105$\times$80$\times$2 image consisting of the
current and the fourth previous pre-processed frame. As in the DQN
experiments~\citep{Mnih15}, we clipped the rewards to be between $-1$
and $+1$, but we did not clip the values of the TD-errors. An episode
started with up to $30$ 'do nothing' actions (\emph{no-op condition})
and it was played until the end of the game or for a maximum of 18,000
frames (i.e., 5 minutes).

We used the meta-parameters in \citet{Elfwing17} as starting values to
initialize the meta-parameters according to Equations~\ref{eq:addN1}
and \ref{eq:addN2}: $\alpha$: 0.001, $\gamma$: 0.99, $\lambda$: 0.8,
$\tau_0$: 0.5, and $\tau_k$: 0.0005. The experiments ran for 2,000
generations (i.e., 200,000 episodes of learning in total). The score
used for selection was the total raw score received by an algorithm
instance in one generation. 

The deep SiL agent in \citet{Elfwing17} outperformed DQN, the Gorila
implementation of DQN~\citep{Nair15} and double DQN~\citep{Hasselt15}
when tested on the 12 games played by DQN that started with the
letters 'A' and 'B'. In this study, we tested the
OMPAC method in three Atari 2600 games: Alien, Amidar, and Assault. We
choose those three because they are games where the deep SiL agents
were outperformed by one or more of the other agents, and it should
therefore be room for significant improvements in performance.

\begin{figure}[!htb]
   \begin{center}
    \subfigure
    {
      \includegraphics[width=0.31\columnwidth]{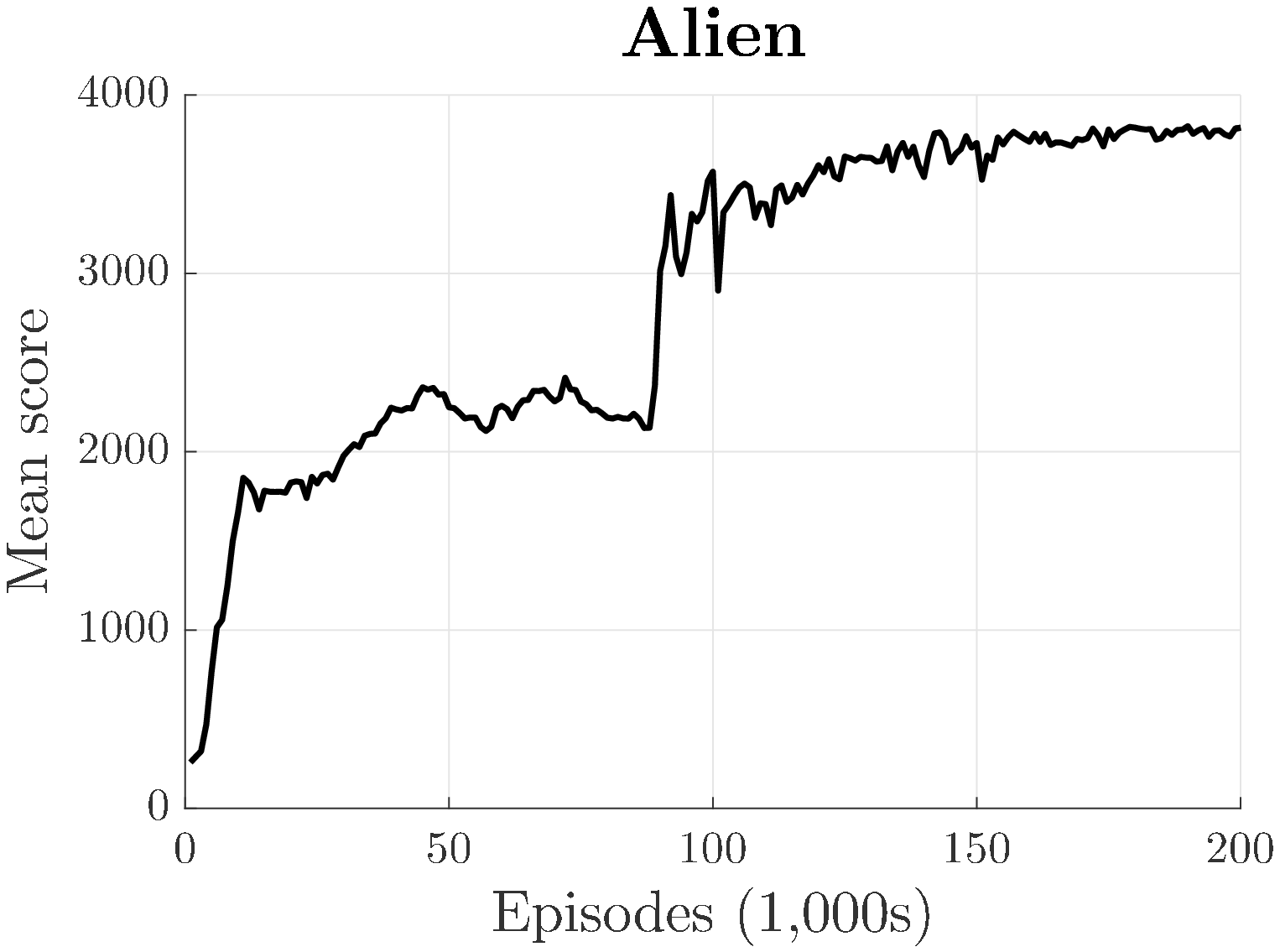} 
    }
    \subfigure
    {
      \includegraphics[width=0.31\columnwidth]{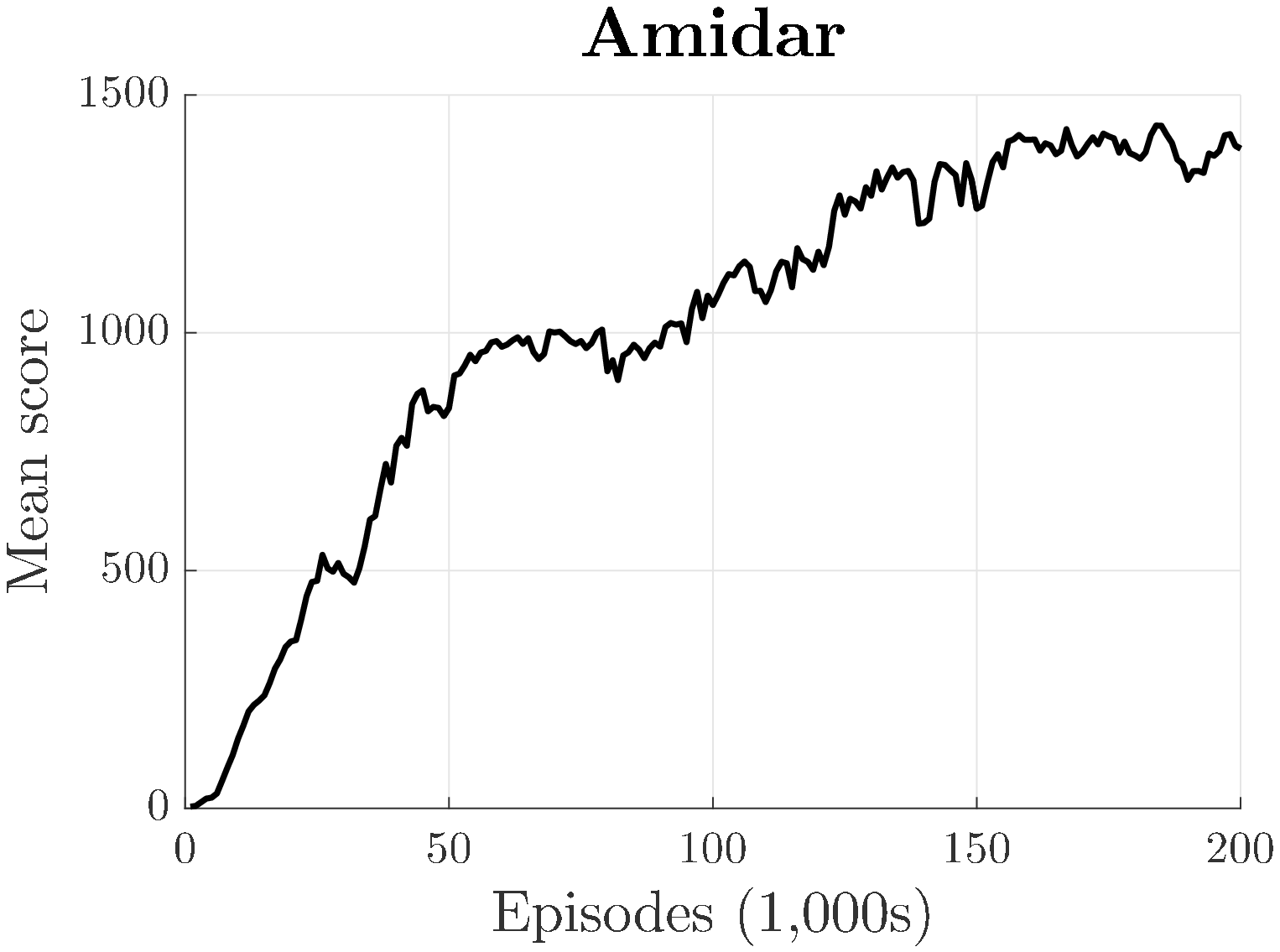} 
    } 
    \subfigure
    {
      \includegraphics[width=0.31\columnwidth]{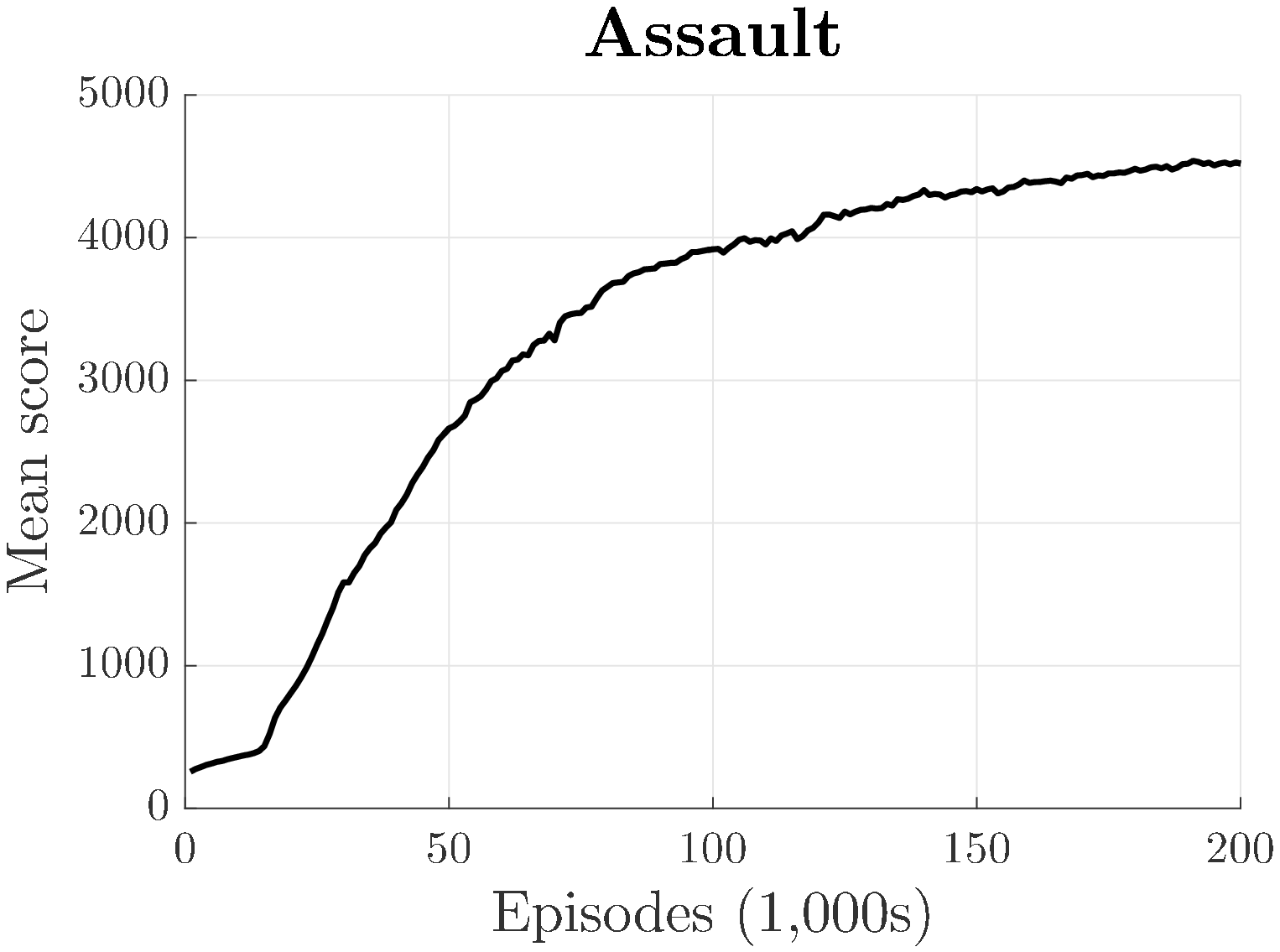} 
    } 
  \end{center}
  \caption{Average learning curves over the 12 algorithm instances in
    the Alien, Amidar, and Assault Atari 2600 games. The figure shows
    mean scores computed over every 10 generations (1,000 episodes).}
  \label{fig:res_atari} 
\end{figure}
Figure~\ref{fig:res_atari} shows the
average learning curves over the 12 algorithm instances for the three
games. Table~\ref{tab:atari_res} summarizes the results as the average scores
in the final generation and the best mean scores of individual
instances in any generation. The table also shows the final and best
mean scores achieved by deep SiL agents without OMPAC adaptation, and
the reported best mean scores achieved by DQN, the Gorila
implementation of DQN, and double DQN. OMPAC adaptation of the
meta-parameters significantly improved the performance of the deep SiL
agents, between 83\,\% and 177\,\% when measured by average final
scores and between 62\,\% and 82\,\% when measured by best mean
scores. The OMPAC scores are also higher than the reported results for
the other three methods, except for the double DQN score in the
Assault game. However, the learned behavior in the Assault was still
good. We tested the best performing agent in last generation using the
final values of the meta-parameters, and it survived for the full 5
minutes in 88 out of 100 episodes. A video of a typical episode
lasting for 5 minutes can be found at 
\url{http://www.cns.atr.jp/~elfwing/videos/assault_OMPAC.mov}.
\begin{table*}[!htb]
\caption{Average scores over the 12 instances in the final generations
  and the best mean scores of individual instances in any generation
  for deep SiL agents with OMPAC adaptation, final and best mean
  scores achieved by deep SiL agents without OMPAC adaptation, and the
  reported best mean scores achieved by DQN, the Gorila implementation
  of DQN, and double DQN in the no-op condition with 5 minutes of
  evaluation time.}
\label{tab:atari_res}
\begin{center}
\begin{tabular}{l|c|c|c|c|c||c|c} 
 {}        &     {}   &    {}       &  {}  & \multicolumn{2}{c||}{\bf{deep SiL}} & \multicolumn{2}{c}{\bf{OMPAC}} \\ 
\bf{Game}  & \bf{DQN} & \bf{Gorila} & \bf{double DQN} &  \bf{Final} & \bf{Best} & \bf{Final} & \bf{Best}\\ 
\hline
\hline
Alien       & 3,069 & 2,621  & 2,907      & 1,370 & 2,246 & 3,798 & \bf{4,097}  \\
Amidar      & 740    &  1,190 & 702         & 762    & 904    & 1,395 & \bf{1,528}  \\
Assault     & 3,359 & 1,450  & \bf{5,023} & 2,415 & 2,944 & 4,529 & 4,769       \\
\end{tabular} 
\end{center}
\end{table*}

\begin{figure}[!htb]
   \begin{center}
      \includegraphics[width=0.47\columnwidth]{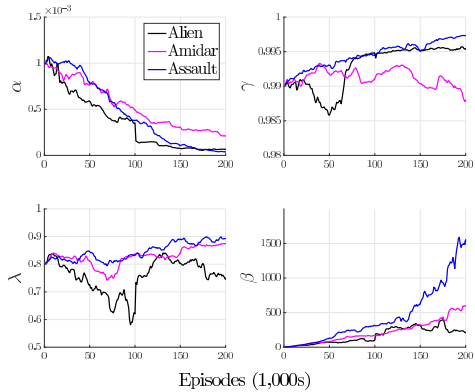} 
  \end{center}
  \caption{Average values of the meta-parameters in the Alien, Amidar,
    and Assault Atari 2600 games for OMPAC adaptation. For
    visualization purposes, instead of showing $\tau(i)$, the figure
    shows the inverse temperature $\beta(i) = 1/\tau(i) =
    (1+\tau_ki)/\tau_0$.}
  \label{fig:meta_atari}
\end{figure}
Figure~\ref{fig:meta_atari} shows average values of the
meta-parameters over the learning process in the three Atari 2600
games. Similar to the two Tetris games, the values of $\alpha$
decreased, by about an order of magnitude of more, over the learning
process. In contrast to the two Tetris games where the values of
$\lambda$ decreased (from a smaller starting value of 0.55), the values
of $\lambda$ either increased to about 0.9 (Amidar and Assault) or
reached about 0.75 after first decreasing to 0.6 (Alien). The
trajectories of the values of $\gamma$ and $\beta$ were different in
the three games: 1) in Assault, both meta-parameters increased over
the whole learning process (final average values of 0.9973 and 1556,
for $\gamma$ and $\beta$, respectively); 2) in Amidar, the value of
$\gamma$ was relatively stable but the value of $\beta$ increased,
but slower than in Assault, over the whole learning process (0.9880
and 599); and 3) in Alien, the value of $\gamma$, similar to
10$\times$10 Tetris, decreased during the first $\sim$50,000 episodes
before it started to increase, and the value of $\beta$, similar to
SZ-Tetris, increased during the first $\sim$120,000 episodes when
learning performance was increasing and it then slowly decreased when
the learning performance plateaued (0.9953 and 206).

\section{Analysis}
In this section, we investigate the ability of the OMPAC method to
learn when starting from bad settings of the meta-parameters. The
experiments in the two Tetris and the Atari 2600 domains showed that
OMPAC adaptation of the meta-parameters can significantly improve the
learning performance when using suitable starting values of the
meta-parameters. However, when, for example, encountering a new task,
it would be valuable to be able to use OMPAC either as a method for
achieving high performance even when the initial values of the
meta-parameters are not suitable for the task at hand or as a method
for finding suitable values of the meta-parameters for future
experiments. 

We chose to limit our investigation to different settings of $\gamma$
and $\tau_k$. We trained shallow dSiL agents in SZ-Tetris with three
settings of the starting value of $\gamma \in \{0.8, 0.9, 0.99\}$ and
three settings of the starting value of $\lambda \in
\{$2.5$\times$10$^{-3}, $ 2.5$\times$10$^{-4}, $
2.5$\times$10$^{-5}\}$. We used the same values of the three other
meta-parameters as in the earlier SZ-Tetris experiment. For each of
the nine settings of $\gamma$ and $\tau_k$, we also trained agents
without OMPAC adaptation for 10 separate runs.

\begin{figure*}[!htb]
   \begin{center}
    \subfigure
    {
      \includegraphics[width=1.0\textwidth]{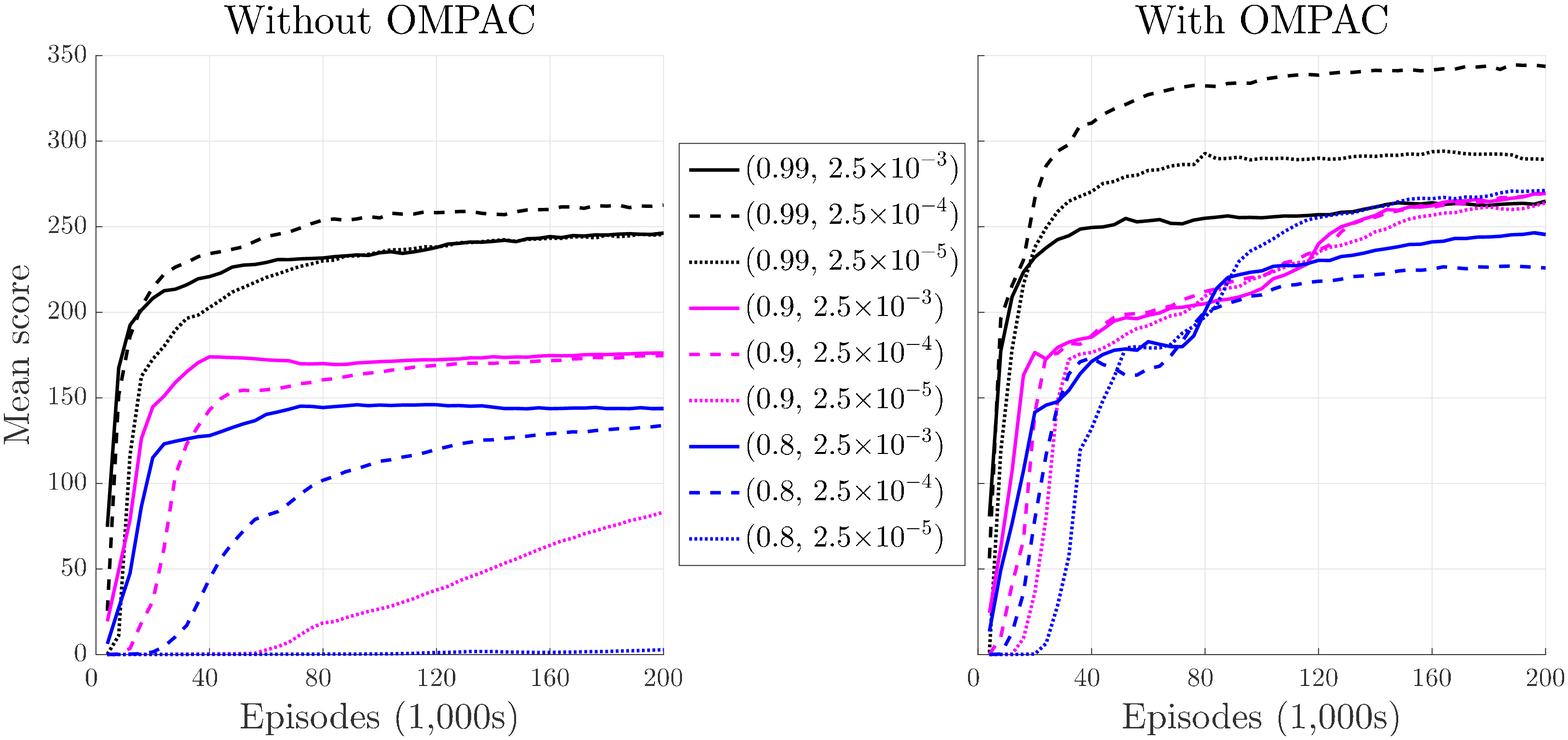} 
    }
    \subfigure
    {
      \includegraphics[width=1.0\textwidth]{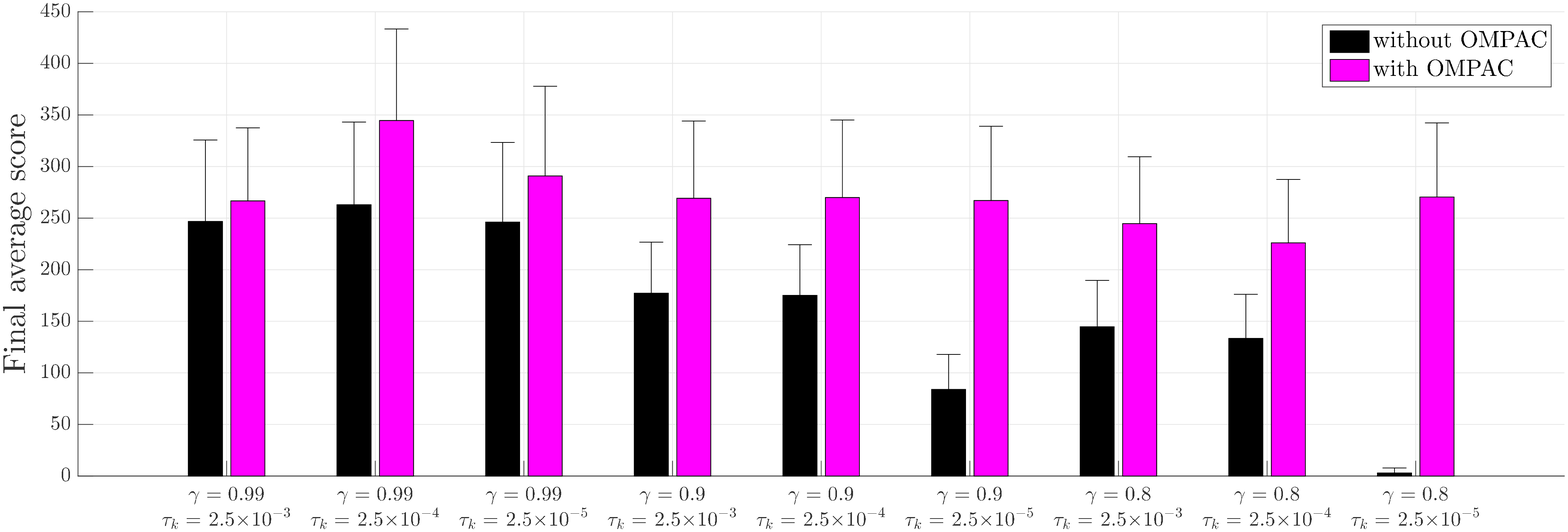} 
    }
  \end{center}
  \caption{The top panel shows average learning curves for shallow
    dSiL agents in SZ-Tetris for different fixed values of $\gamma$
    and $\tau_k$ (left) and for different starting values of the
    same meta-parameters for OMPAC adaptation (right). The three
    colors corresponds to different values of $\gamma$: black for
    0.99, purple for 0.9, and blue for 0.8. The three line styles
    corresponds to different values of $\tau_k$: solid for
    2.5$\times$10$^{-3}$, dashed for 2.5$\times$10$^{-4}$, dotted for
    2.5$\times$10$^{-5}$. The bottom panel shows average scores over
    the final 1000 episodes with standard deviations}
  \label{fig:test_ompac}
\end{figure*}
Figure~\ref{fig:test_ompac} shows average learning curves with and
without OMPAC adaptation of the meta-parameters (top panel), and the
average scores over the final 1000 episodes with standard deviations
(bottom panel). The results clearly show that the OMPAC method is able
to overcome bad initial settings of meta-parameters. The two largest
improvements in average final scores (267 and 183 points) were for the
$(\gamma, \lambda)$-settings ((0.8, 2.5$\times$10$^{-5}$) and (0.9,
2.5$\times$10$^{-5}$)) that achieved the lowest average final scores
(3 and 84 points) without OMPAC adaptation.

\section{Conclusions}
In this study, we proposed the OMPAC method for online adaptation of
the meta-parameters in reinforcement learning by competition between
algorithm instances running in parallel. We validated the proposed
method by significantly improving the state-of-the-art scores in
stochastic SZ-Tetris and 10$\times$10 Tetris, and by significantly
improving the performance of deep Sarsa($\lambda$) agents in three
Atari 2600 games. The experiments also demonstrated the ability of the
OMPAC method to adapt the meta-parameters according to the learning
progress in different tasks. 

\section*{Acknowledgments}
This work was supported by the project commissioned by the New Energy
and Industrial Technology Development Organization (NEDO), JSPS
KAKENHI grant 16K12504, and Okinawa Institute of Science and
Technology Graduate University research support to KD.

\bibliography{learning}
\bibliographystyle{apalike}

\end{document}